\documentclass[preprint,review,12pt]{elsarticle}

\usepackage{amssymb}
\usepackage{epstopdf}
\usepackage{subfig}
\usepackage{graphicx}
\usepackage{amsmath,amssymb} % define this before the line numbering.
\usepackage{upgreek}
\usepackage{color}
\usepackage[usenames,dvipsnames,table]{xcolor}% http://ctan.org/pkg/xcolor
\usepackage[pagebackref=true,breaklinks=true,letterpaper=true,colorlinks,bookmarks=false]{hyperref}
\usepackage{enumitem}
\usepackage{tikz}
\usepackage{tabularx,colortbl}
\usepackage{booktabs}

\setlist{noitemsep}
\newcommand{\etal}{\textit{et al.}}

\journal{Pattern Recognition}

\begin{document}

\begin{frontmatter}



%\title{End-to-end scene text understanding in multi-lingual environments}
\title{Improving patch-based scene text script identification with ensembles of conjoined networks}


\author{Lluis~Gomez}
\author{Anguelos~Nicolaou}
\author{Dimosthenis~Karatzas}
\address{Computer Vision Center, Universitat Autonoma de Barcelona. Edifici O, Campus UAB, 08193 Bellaterra (Cerdanyola) Barcelona, Spain. E-mail: {lgomez,dimos}@cvc.uab.cat}

\begin{abstract}
This paper focuses on the problem of script identification in scene text images. Facing this problem with state of the art CNN classifiers is not straightforward, as they fail to address a key characteristic of scene text instances: their extremely variable aspect ratio. Instead of resizing input images to a fixed aspect ratio as in the typical use of holistic CNN classifiers, we propose here a patch-based classification framework in order to preserve discriminative parts of the image that are characteristic of its class.

We describe a novel method based on the use of ensembles of conjoined networks to jointly learn discriminative stroke-parts representations and their relative importance in a patch-based classification scheme. Our experiments with this learning procedure demonstrate state-of-the-art results in two public script identification datasets.

In addition, we propose a new public benchmark dataset for the evaluation of multi-lingual scene text end-to-end reading systems. Experiments done in this dataset demonstrate the key role of script identification in a complete end-to-end system that combines our script identification method with a previously published text detector and an off-the-shelf OCR engine.

\end{abstract}

\begin{keyword}
script identification \sep scene text understanding \sep multi-language OCR \sep convolutional neural networks \sep ensemble of conjoined networks
\end{keyword}

\end{frontmatter}



%%%%%%%%%%%%%%%%%%%%%%%%%%%%%%%%%%%%%%%%%%%%%%%%%%%%%%%%%%%%%
\section{Introduction}
\label{sec:intro}  % \label{} allows reference to this section
Script and language identification are important steps in modern OCR systems designed for multi-language environments. Since text recognition algorithms are language-dependent, detecting the script and language at hand allows selecting the correct language model to employ~\cite{unnikrishnan2009}. While script identification has been widely studied in document analysis~\cite{ghosh2010script,pal2004indian}, it remains an almost unexplored problem for scene text. In contrast to document images, scene text presents a set of specific challenges, stemming from the high variability in terms of perspective distortion, physical appearance, variable illumination and typeface design. At the same time, scene text comprises typically a few words, contrary to longer text passages available in document images.

Current end-to-end systems for scene text reading~\cite{bissacco2013,Jaderberg2014,Neumann2015} assume single script and language inputs given beforehand, i.e. provided by the user, or inferred from available meta-data. % (e.g. geo-tagging, media provenance, etc.). 
The unconstrained text understanding problem for large collections of images from unknown sources has not been considered up to very recently~\cite{Shi2015,Shi2016,Nicolaou2016,Gomez2016,Tian2016125}.
While there exists some previous research in script identification of text over complex backgrounds~\cite{gllavata2005script,shivakumara2015}, such methods have been so far limited to video overlaid-text, which presents in general different challenges than scene text.
%DK_11/5: I have removed the following paragraph, since the focus is only on script identification.
%On the contrary, the problem of multi-script text detection in natural scenes has received some attention from the research community~\cite{Pan2009,kasar2012,yao2012detecting,Gomez2013}. It should be noted nevertheless that such methods are typically evaluated on datasets with very limited script variability: E.g. Latin and Chinese in~\cite{Pan2009}, Latin and Kannada in~\cite{kumar2013multi}, or Latin and Hangul in~\cite{lee2010}. %Script identification, can be seen as a fine-grained classification of detected text into script classes. 

\begin{figure}[t]
\includegraphics[width=\linewidth]{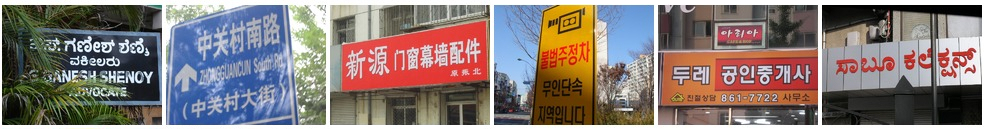}
\caption{Collections of images from unknown sources may contain textual information in different scripts.}
\label{fig:babel_db}
\end{figure}

\begin{figure}[h]
\includegraphics[width=\linewidth]{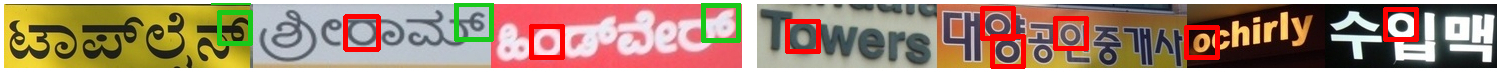}
\caption{(best viewed in color) Certain stroke-parts (in green) are discriminative for the identification of a particular script (left), while others (in red) can be trivially ignored because are frequent in other classes (right).}
\label{fig:stroke_parts}
\end{figure}

This paper addresses the problem of %joint text detection and 
script identification in natural scene images, paving the road towards true multi-lingual end-to-end scene text understanding.
Multi-script text exhibits high intra-class variability (words written in the same script vary a lot) and high inter-class similarity (certain scripts resemble each other). Examining text samples from different scripts, it is clear that some stroke-parts are quite discriminative, whereas others can be trivially ignored as they occur in multiple scripts. The ability to distinguish these relevant stroke-parts can be leveraged for recognising the corresponding script. Figure~\ref{fig:stroke_parts} shows an example of this idea.

\begin{figure}[h]
\includegraphics[width=\linewidth]{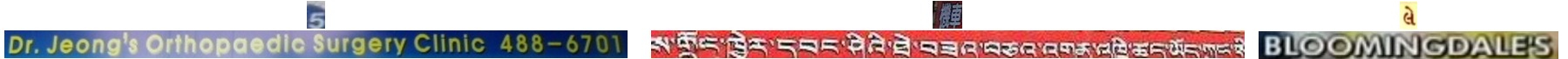}
\caption{Scene text images with the larger/smaller aspect ratio available in three different datasets: MLe2e(left), SIW-13(center), and CVSI(right).}
\label{fig:aspect_ratio_variability}
\end{figure}

The use of state of the art CNN classifiers for script identification is not straightforward, as they fail to address a key characteristic of scene text instances: their extremely variable aspect ratio.  As can be seen in Figure~\ref{fig:aspect_ratio_variability}, scene text images may span from single characters to long text sentences, and thus resizing images to a fixed aspect ratio, as in the typical use of holistic CNN classifiers, will deteriorate discriminative parts of the image that are characteristic of its class. The key intuition behind the proposed method is that in order to retain the discriminative power of stroke parts we must rely in powerful local feature representations and use them within a patch-based classifier. In other words, while holistic CNNs have superseded patch-based methods for image classification, we claim that patch-based classifiers can still be essential in tasks where image shrinkage is not feasible.

In previously published work~\cite{Gomez2016} we have presented a method combining convolutional features, extracted by sliding a window with a single layer Convolutional Neural Network (CNN)~\cite{coates2011analysis}, and the Naive-Bayes Nearest Neighbour (NBNN) classifier~\cite{boiman2008defense} with promising results. In this paper we demonstrate far superior performance by extending our previous work in two different ways: First, we use deep CNN architectures in order to learn more discriminative representations for the individual image patches; Second, we propose a novel learning methodology to jointly learn the patch representations and their importance (contribution) in a global image to class probabilistic measure. For this, we train our CNN using an Ensemble of Conjoined Networks and a loss function that takes into account the global classification error for a group of $N$ patches instead of looking only into a single image patch. Thus, at training time our network is presented with a group of $N$ patches sharing the same class label and produces a single probability distribution over the classes for all them. This way we model the goal for which the network is trained, not only to learn good local patch representations, but also to learn their relative importance in the global image classification task.

Experiments performed over two public datasets for scene text classification demonstrate state-of-the-art results. In particular we are able to reduce classification error by $5$ percentage points in the SIW-13 dataset. We also introduce a new benchmark dataset, namely the MLe2e dataset, for the evaluation of scene text end-to-end reading systems and all intermediate stages such as text detection, script identification and text recognition.
%for the evaluation of joint text detection and script identification in natural scenes. 
The dataset contains a total of $711$ scene images, and $1821$ text line instances, covering four different scripts (Latin, Chinese, Kannada, and Hangul) and a large variability of scene text samples.

%------------------------------------------------------------------------- 
\section{Related Work}
\label{sec:background}
Script identification is a well studied problem in document image analysis. Gosh~\etal~\cite{ghosh2010script} has published a compehensive review of methods dealing with this problem. They identify two broad categories of methods: structure-based and visual appearance-based techniques. In the first category, Spitz and Ozaki~\cite{spitz1995palace,spitz1997determination} propose the use of the vertical distribution of upward concavities in connected components and their optical density for page-wise script identification. Lee~\etal~\cite{lee1998language}, and Waked~\etal~\cite{waked1998skew} among others build on top of Spitz seminal work by incorporating additional connected component based features. Similarly, Chaudhuri~\etal~\cite{chaudhury1999trainable} use the projection profile, statistical and topological features, and stroke features for classification of text lines in printed documents. Hochberg~\etal~\cite{judith1995automatic} propose the use of cluster-based templates to identify unique characteristic shapes. A method that is similar in spirit with the one presented in this paper, while requiring textual symbols to be precisely segmented to generate the templates.

Regarding segmentation-free methods based on visual appearance of scripts, i.e. not directly analyzing the character patterns in the document, Wood~\etal~\cite{wood1995language} experimented with the use of vertical and horizontal projection profiles of full-page document images. More recent methods in this category have used texture features from Gabor filters analysis~\cite{tan1998rotation,Chan20012523,pan2005script} or Local Binary Patterns~\cite{ferrer2013lbp}. Neural networks have been also used for segmentation-free script identification~\etal~\cite{Jain1996743,chi2003hierarchical} without the use of hand-crafted features.

All the methods discussed above are designed specifically with printed document images in mind. Structure-based methods require text connected components to be precisely segmented from the image, while visual appearance-based techniques are known to work better in bilevel text. Moreover, some of these methods require large blocks of text in order to obtain sufficient information and thus are not well suited for scene text which typically comprises a few words.

Contrary to the case of printed document images, research in script identification on non traditional paper layouts is more scarce, and has been mainly dedicated to handwritten text~\cite{Hennig2002445,Schenk20093383,Zhu20093184,Basu20103507,Zhong20151211}, and video overlaid-text~\cite{gllavata2005script,sharma2013word,phan2011video,shivakumara2014gradient,shivakumara2015} until very recently. Gllavatta \etal~\cite{gllavata2005script}, in the first work dealing with video text script identification, proposed a method using the wavelet transform to detect edges in overlaid-text images. Then, they extract a set of low-level edge features, and make use of a K-NN classifier.

Sharma \etal~\cite{sharma2013word} have explored the use of traditional document analysis techniques for video overlaid-text script identification at word level. They analyze three sets of features: Zernike moments, Gabor filters, and a set of hand-crafted gradient features previously used for handwritten character recognition. They propose a number of pre-processing algorithms to overcome the inherent challenges of video overlaid-text. In their experiments the combination of super resolution, gradient features, and a SVM classifier perform significantly better that the other combinations.

Phan \etal~\cite{phan2011video} propose a method for combined detection of video text overlay and script identification. They propose the extraction of upper and lower extreme points for each connected component of Canny edges of text lines and analyse their smoothness and cursiveness.

Shivakumara \etal~\cite{shivakumara2014gradient,shivakumara2015} rely on skeletonization of the dominant gradients They analyze the angular curvatures~\cite{shivakumara2014gradient} of skeleton components, and the spatial/structural~\cite{shivakumara2015} distribution of their end, joint, and intersection points to extract a set of hand-crafted features. For classification they build a set of feature templates from train data, and use the Nearest Neighbor rule for classifying scripts at word~\cite{shivakumara2014gradient} or text block~\cite{shivakumara2015} level.

As said before, all these methods have been designed (and evaluated) specifically for video overlaid-text, which presents in general different challenges than scene text. Concretely, they mainly rely in accurate edge detection of text components and this is not always feasible in scene text. 

More recently, Sharma \etal~\cite{sharma2015bag} explored the use of Bag-of-Visual Words based techniques for word-wise script identification in video-overlaid text. They use Bag-Of-Features (BoF) and Spatial Pyramid Matching (SPM) with patch based SIFT descriptors and found that the SPM pipeline outperforms traditional script identification techniques involving gradient based features (e.g. HoG) and texture based features (e.g. LBP).

In 2015, the ICDAR Competition on Video Script Identification (CVSI-2015)~\cite{Sharmai2015} challenged the document analysis community with a new competitive benchmark dataset. With images extracted from different video sources (news, sports etc.) covering mainly overlaid-text, but also a few instances of scene text. The top performing methods in the competition where all based in Convolutional Neural Networks, showing a clear difference in overall accuracy over pipelines using hand-crafted features (e.g. LBP and/or HoG).

The first dataset for script identification in real scene text images was provided by Shi~\etal in~\cite{Shi2015}, where the authors propose the Multi-stage Spatially-sensitive Pooling Network (MSPN) method. The MSPN network overcomes the limitation of having a fixed size input in traditional Convolutional Neural Networks by pooling along each row of the intermediate layers' outputs by taking the maximum (or average) value in each row. Their method is extended in~\cite{Shi2016} by combining deep features and mid-level representations into a globally trainable deep model. They extract local deep features at every layer of the MSPN and describe images with a codebook-based encoding method that can be used to fine-tune the CNN weights.

Nicolaou~\etal~\cite{Nicolaou2016} has presented a method based on texture features producing state of the art results in script identification for both scene or overlaid text images. They rely in hand-crafted texture features, a variant of LBP, and a deep Multi Layer Perceptron to learn a metric space in which they perform K-NN classification.

In our previous work~\cite{Gomez2016} we have proposed a patch-based method for script identification in scene text images. We used Convolutional features, extracted from small image patches, and the Naive-Bayes Nearest Neighbour classifier (NBNN). We also presented a simple weighting strategy in order to discover the most discriminative parts (or templates patches) per class in a fine-grained classification approach.

In this paper we build upon our previous work~\cite{Gomez2016} by extending it in two ways: On one side, we make use of a much deeper Convolutional Neural Network model. On the other hand, we replace the weighted NBNN classifier by a patch-based classification rule that can be integrated in the CNN training process by using an Ensemble of Conjoined Networks. This way, our CNN model is able to learn at the same time expressive representations for image patches and their relative contribution to the patch-based classification rule.

From all reviewed methods the one proposed here is the only one based in a patch-based classification framework. Our intuition is that in cases where holistic CNN models are not directly applicable, as in the case of text images (because of their highly variable aspect ratios), the contribution of rich parts descriptors without any deterioration (either by image distortion or by descriptor quantization) is essential for correct image classification. 

In this sense our method is related with some CNN extensions that have been proposed for video classification.  Unlike still images which can be cropped and rescaled to a fixed size, video sequences have a variable temporal dimension and cannot be directly processed with a fixed-size architecture. In this context, 3D Convolutional Neural Networks~\cite{baccouche2011sequential,ji20133d} have been proposed to leverage the motion information encoded in multiple contiguous frames. Basically the idea is to feed the CNN with a stack of a fixed number of consecutive frames and perform convolutions in both time and space dimensions. Still these methods require a fixed size input and thus they must be applied several times through the whole sequence to obtain a chain of outputs that are then averaged~\cite{ji20133d} or fed into an Recurrent Neural Network~\cite{baccouche2011sequential} to provide a final decision. Karpathy \etal~\cite{karpathy2014large} also treat videos as bags of short fixed-length clips, but they investigate the use of different temporal connectivity patterns (early fusion, late fusion and slow fusion). To produce predictions for an entire video they randomly sample 20 clips and take the average of the network class predictions. While we share with these methods the high-level goal of learning CNN weights from groups of stacked patches (or frames) there are two key differences in the way we build our framework: (1) the groups of patches that are fed into the network at training time are randomly sampled and do not follow any particular order; and (2) at test time we decouple the network to densely evaluate single patches and average their outputs. In other words, while in stacked-frame CNNs for video recognition having an ordered sequence of input patches is crucial to learn spatio-temporal features, our design aims to learn which are the most discriminative patches in the input stack, independently of their relative spatial arrangement.

In the experimental section we compare our method with some of the algorithms reviewed in this section and demonstrate its superiority. Concretely our approach improves the state-of-the-art in the SIW-13~\cite{Shi2016} dataset for scene text script classification by a large margin of $5$ percentage points, while performs competitively in the CVSI-2015~\cite{Sharmai2015} video overlaid-text dataset.

%This is maybe interesting to add value in the description of the MLe2e dataset.

%\textcolor{red}{The problem of multi-script text detection in natural scenes has received the attention of the research community in recent years~\cite{Pan2009,kasar2012,yao2012detecting,Gomez2013}, although such methods are always evaluated in datasets with very limited script variability: E.g. Latin and Chinese in~\cite{Pan2009}, Latin and Kannada in~\cite{kumar2013multi}, or Latin and Hangul in~\cite{lee2010}.}

%------------------------------------------------------------------------- 
\section{Patch-based classification with Ensembles of Conjoined Networks}
\label{sec:method}
In our patch-based classification method an input image is represented as a collection of local descriptors, from patches extracted following a certain sampling strategy. Those local features are then fed into a global classifier rule, that makes a decision for the input image.

\subsection{Convolutional Neural Network for image-patch classification}
\label{sec:simple_cnn}

\begin{figure*}[b]
\centering
\subfloat[]{\includegraphics[width=0.3\linewidth]{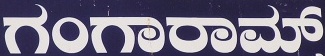}}
\hspace{1em}
\subfloat[]{\includegraphics[width=0.17\linewidth]{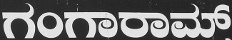}}\\
\subfloat[]{\includegraphics[width=\linewidth]{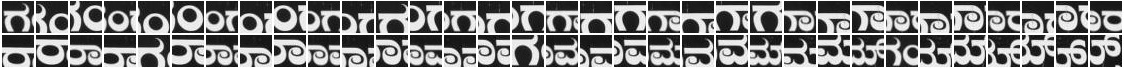}}
\caption{The original scene text images (a) are converted to greyscale and resized to a fixed height (b) in order to extract small local patches with a dense sampling strategy (c).}
\label{fig:sampling}
\end{figure*}

Given an input scene text image (i.e. a pre-segmented word or text line) we first resize it to a fixed height of $40$ pixels, but retaining its original aspect ratio. Since scene text can appear in any possible combination of foreground and background colors, we pre-process the image by converting it into grayscale and centering pixel values. Then, we densely extract patches at two different scales, $32\times32$ and $40\times40$, by sliding a window with a step of $8$ pixels. The particular values of these two window scales and step size was found by cross-validation optimization as explained in section~\ref{sec:implementation}, and its choice can be justified as follows: the $40\times40$ patch, covering the full height of the resized image, is a natural choice in our system because it provides the largest squared region we can crop; the $32\times32$ patches are conceived for better scale invariance of the CNN model, similarly as the random crops typically used for data augmentation in CNN-based image classification~\cite{Krizhevsky2012}. Figure~\ref{fig:sampling} shows the patches extracted from a given example image. This way we build a large dataset of image patches that take the same label as the image they were extracted from. With this dataset of patches we train a CNN classifier for the task of individual image patch classification. 

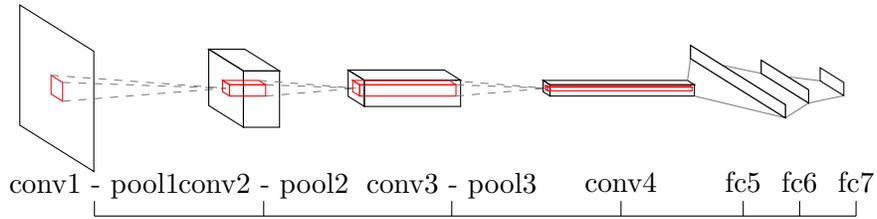
\begin{figure*}[h]
\centering
\begin{tikzpicture}
\pgfmathsetmacro{\layeroffset}{1.25}

% ====================  Conv 1 layer ================ %
\pgfmathsetmacro{\offsetx}{-1}
\pgfmathsetmacro{\channels}{0}
\pgfmathsetmacro{\mapsize}{32/40}
\pgfmathsetmacro{\kernelsize}{5/40}
%draw map block
\draw[black] (\offsetx+\mapsize,\mapsize,\mapsize) -- ++(-2*\mapsize,0,-2*\mapsize) -- ++(0,-2*\mapsize,0) -- ++(2*\mapsize,0,2*\mapsize) -- cycle;
%draw kernel block
\draw[red] (\offsetx+\kernelsize,\kernelsize,\kernelsize) -- ++(-2*\kernelsize,0,-2*\kernelsize) -- ++(0,-2*\kernelsize,0) -- ++(2*\kernelsize,0,2*\kernelsize) -- cycle;

% ====================  Pool 1 layer ================ %
\pgfmathsetmacro{\offsetxlast}{\offsetx+\channels}
\pgfmathsetmacro{\offsetx}{\channels+\layeroffset}
%draw pooling piramid
\draw[dashed,gray] (\offsetxlast+\kernelsize,\kernelsize,\kernelsize) -- (\offsetx,0,0);
\draw[dashed,gray] (\offsetxlast+\kernelsize,-\kernelsize,\kernelsize) -- (\offsetx,0,0);
\draw[dashed,gray] (\offsetxlast-\kernelsize,\kernelsize,-\kernelsize) -- (\offsetx,0,0);

% ====================  Conv 2 layer ================ %
\pgfmathsetmacro{\channels}{96/200}
\pgfmathsetmacro{\mapsize}{15/40}
\pgfmathsetmacro{\kernelsize}{3/40}
%draw map block
\draw[black] (\offsetx+\mapsize,\mapsize,\mapsize) -- ++(-2*\mapsize,0,-2*\mapsize) -- ++(0,-2*\mapsize,0) -- ++(2*\mapsize,0,2*\mapsize) -- cycle;
\draw[black] (\offsetx+\mapsize,\mapsize,\mapsize) -- ++(\channels,0,0) -- ++(0,-2*\mapsize,0) -- ++(-\channels,0,0) -- cycle;
\draw[black] (\offsetx+\mapsize,\mapsize,\mapsize) -- ++(\channels,0,0) -- ++(-2*\mapsize,0,-2*\mapsize) -- ++(-\channels,0,0) -- cycle;
%draw kernel block
\draw[red] (\offsetx+\kernelsize,\kernelsize,\kernelsize) -- ++(-2*\kernelsize,0,-2*\kernelsize) -- ++(0,-2*\kernelsize,0) -- ++(2*\kernelsize,0,2*\kernelsize) -- cycle;
\draw[red] (\offsetx+\kernelsize,\kernelsize,\kernelsize) -- ++(\channels,0,0) -- ++(0,-2*\kernelsize,0) -- ++(-\channels,0,0) -- cycle;
\draw[red] (\offsetx+\kernelsize,\kernelsize,\kernelsize) -- ++(\channels,0,0) -- ++(-2*\kernelsize,0,-2*\kernelsize) -- ++(-\channels,0,0) -- cycle;

% ====================  Pool 2 layer ================ %
\pgfmathsetmacro{\offsetxlast}{\offsetx+\channels}
\pgfmathsetmacro{\offsetx}{\offsetx+\channels+\layeroffset}
%draw pooling piramid
\draw[dashed,gray] (\offsetxlast+\kernelsize,\kernelsize,\kernelsize) -- (\offsetx,0,0);
\draw[dashed,gray] (\offsetxlast+\kernelsize,-\kernelsize,\kernelsize) -- (\offsetx,0,0);
\draw[dashed,gray] (\offsetxlast-\kernelsize,\kernelsize,-\kernelsize) -- (\offsetx,0,0);

% ====================  Conv 3 layer ================ %
\pgfmathsetmacro{\channels}{256/200}
\pgfmathsetmacro{\mapsize}{7/40}
\pgfmathsetmacro{\kernelsize}{3/40}
%draw map block
\draw[black] (\offsetx+\mapsize,\mapsize,\mapsize) -- ++(-2*\mapsize,0,-2*\mapsize) -- ++(0,-2*\mapsize,0) -- ++(2*\mapsize,0,2*\mapsize) -- cycle;
\draw[black] (\offsetx+\mapsize,\mapsize,\mapsize) -- ++(\channels,0,0) -- ++(0,-2*\mapsize,0) -- ++(-\channels,0,0) -- cycle;
\draw[black] (\offsetx+\mapsize,\mapsize,\mapsize) -- ++(\channels,0,0) -- ++(-2*\mapsize,0,-2*\mapsize) -- ++(-\channels,0,0) -- cycle;
%draw kernel block
\draw[red] (\offsetx+\kernelsize,\kernelsize,\kernelsize) -- ++(-2*\kernelsize,0,-2*\kernelsize) -- ++(0,-2*\kernelsize,0) -- ++(2*\kernelsize,0,2*\kernelsize) -- cycle;
\draw[red] (\offsetx+\kernelsize,\kernelsize,\kernelsize) -- ++(\channels,0,0) -- ++(0,-2*\kernelsize,0) -- ++(-\channels,0,0) -- cycle;
\draw[red] (\offsetx+\kernelsize,\kernelsize,\kernelsize) -- ++(\channels,0,0) -- ++(-2*\kernelsize,0,-2*\kernelsize) -- ++(-\channels,0,0) -- cycle;

% ====================  Pool 3 layer ================ %
\pgfmathsetmacro{\offsetxlast}{\offsetx+\channels}
\pgfmathsetmacro{\offsetx}{\offsetx+\channels+\layeroffset}
%draw pooling piramid
\draw[dashed,gray] (\offsetxlast+\kernelsize,\kernelsize,\kernelsize) -- (\offsetx,0,0);
\draw[dashed,gray] (\offsetxlast+\kernelsize,-\kernelsize,\kernelsize) -- (\offsetx,0,0);
\draw[dashed,gray] (\offsetxlast-\kernelsize,\kernelsize,-\kernelsize) -- (\offsetx,0,0);

% ====================  Conv 4 layer ================ %
\pgfmathsetmacro{\channels}{384/200}
\pgfmathsetmacro{\mapsize}{3/40}
\pgfmathsetmacro{\kernelsize}{1/40}
%draw map block
\draw[black] (\offsetx+\mapsize,\mapsize,\mapsize) -- ++(-2*\mapsize,0,-2*\mapsize) -- ++(0,-2*\mapsize,0) -- ++(2*\mapsize,0,2*\mapsize) -- cycle;
\draw[black] (\offsetx+\mapsize,\mapsize,\mapsize) -- ++(\channels,0,0) -- ++(0,-2*\mapsize,0) -- ++(-\channels,0,0) -- cycle;
\draw[black] (\offsetx+\mapsize,\mapsize,\mapsize) -- ++(\channels,0,0) -- ++(-2*\mapsize,0,-2*\mapsize) -- ++(-\channels,0,0) -- cycle;
%draw kernel block
\draw[red] (\offsetx+\kernelsize,\kernelsize,\kernelsize) -- ++(-2*\kernelsize,0,-2*\kernelsize) -- ++(0,-2*\kernelsize,0) -- ++(2*\kernelsize,0,2*\kernelsize) -- cycle;
\draw[red] (\offsetx+\kernelsize,\kernelsize,\kernelsize) -- ++(\channels,0,0) -- ++(0,-2*\kernelsize,0) -- ++(-\channels,0,0) -- cycle;
\draw[red] (\offsetx+\kernelsize,\kernelsize,\kernelsize) -- ++(\channels,0,0) -- ++(-2*\kernelsize,0,-2*\kernelsize) -- ++(-\channels,0,0) -- cycle;

% ====================  FC 5 layer ================ %
\pgfmathsetmacro{\offsetxlast}{\offsetx+\channels}
\pgfmathsetmacro{\offsetx}{\offsetx+\channels+\layeroffset/2}
\pgfmathsetmacro{\lastmapsize}{\mapsize}
\pgfmathsetmacro{\mapsize}{7/40}
\pgfmathsetmacro{\channels}{4096/4000}
%draw full connections
\draw[gray] (\offsetxlast+\lastmapsize,-\lastmapsize,\lastmapsize) -- (\offsetx+\channels,0,\channels);
\draw[gray] (\offsetxlast-\lastmapsize,\lastmapsize,-\lastmapsize) -- (\offsetx+\channels-2*\channels,0,-\channels);
%draw fc plane
\draw[black,fill=white] (\offsetx+\channels,0,\channels) -- ++(0,\mapsize,0) -- ++(-2*\channels,0,-2*\channels) -- ++(0,-\mapsize,0) -- cycle;
\coordinate (fc5x1) at (\offsetx+\channels,0,\channels);
\coordinate (fc5x2) at (0,\mapsize,0);
\coordinate (fc5x3) at (-2*\channels,0,-2*\channels);
\coordinate (fc5x4) at (0,-\mapsize,0);

% ====================  FC 6 layer ================ %
\pgfmathsetmacro{\offsetxlast}{\offsetx}
\pgfmathsetmacro{\offsetx}{\offsetx+\layeroffset/2}
\pgfmathsetmacro{\mapsize}{7/40}
\pgfmathsetmacro{\channelslast}{\channels}
\pgfmathsetmacro{\channels}{2048/4000}
%draw full connections
\draw[gray] (\offsetxlast+\channelslast,0,\channelslast) -- (\offsetx+\channels,0,\channels);
\draw[gray] (\offsetxlast+\channelslast-2*\channelslast,0,-\channelslast) -- (\offsetx+\channels-2*\channels,0,-\channels);
%draw fc plane
\draw[black,fill=white] (\offsetx+\channels,0,\channels) -- ++(0,\mapsize,0) -- ++(-2*\channels,0,-2*\channels) -- ++(0,-\mapsize,0) -- cycle;
\coordinate (fc6x1) at (\offsetx+\channels,0,\channels);
\coordinate (fc6x2) at (0,\mapsize,0);
\coordinate (fc6x3) at (-2*\channels,0,-2*\channels);
\coordinate (fc6x4) at (0,-\mapsize,0);

% ====================  FC 7 layer ================ %
\pgfmathsetmacro{\offsetxlast}{\offsetx}
\pgfmathsetmacro{\offsetx}{\offsetx+\layeroffset/2}
\pgfmathsetmacro{\mapsize}{7/40}
\pgfmathsetmacro{\channelslast}{\channels}
\pgfmathsetmacro{\channels}{1024/4000}
%draw full connections
\draw[gray] (\offsetxlast+\channelslast,0,\channelslast) -- (\offsetx+\channels,0,\channels);
\draw[gray] (\offsetxlast+\channelslast-2*\channelslast,0,-\channelslast) -- (\offsetx+\channels-2*\channels,0,-\channels);
%draw fc plane
\draw[black,fill=white] (\offsetx+\channels,0,\channels) -- ++(0,\mapsize,0) -- ++(-2*\channels,0,-2*\channels) -- ++(0,-\mapsize,0) -- cycle;

% draw FC layers again to hide fc connections
\draw[black,fill=white] (fc5x1) -- ++(fc5x2) -- ++(fc5x3) -- ++(fc5x4) -- cycle;
\draw[black,fill=white] (fc6x1) -- ++(fc6x2) -- ++(fc6x3) -- ++(fc6x4) -- cycle;

% draw text
\draw[black] (-0.5,-1.7) -- (7.7*\layeroffset,-1.7);
\node[] at (-0.5,-1.3) {\small conv1 - pool1};
\draw[black] (-0.5,-1.5) -- (-0.5,-1.7);
\node[] at (1.4*\layeroffset,-1.3) {\small conv2 - pool2};
\draw[black] (1.4*\layeroffset,-1.5) -- (1.4*\layeroffset,-1.7);
\node[] at (3.4*\layeroffset,-1.3) {\small conv3 - pool3};
\draw[black] (3.4*\layeroffset,-1.5) -- (3.4*\layeroffset,-1.7);
\node[] at (5.2*\layeroffset,-1.25) {\small conv4};
\draw[black] (5.2*\layeroffset,-1.5) -- (5.2*\layeroffset,-1.7);
\node[] at (6.5*\layeroffset,-1.25) {\small fc5};
\draw[black] (6.5*\layeroffset,-1.5) -- (6.5*\layeroffset,-1.7);
\node[] at (7.1*\layeroffset,-1.25) {\small fc6};
\draw[black] (7.1*\layeroffset,-1.5) -- (7.1*\layeroffset,-1.7);
\node[] at (7.7*\layeroffset,-1.25) {\small fc7};
\draw[black] (7.7*\layeroffset,-1.5) -- (7.7*\layeroffset,-1.7);
\end{tikzpicture}
\caption{Network architecture of the CNN trained to classify individual image patches. The network has three convolutional+pooling stages followed by an extra convolution and three fully connected layers.}
\label{fig:simple_cnn}
\end{figure*}

We use a Deep Convolutional Neural Network to build the expressive image patch representations needed in our method. For the design of our network we start from the CNN architecture proposed in~\cite{Shi2015} as it is known to work well for script identification. We then iteratively do an exhaustive search to optimize by cross-validation the following CNN hyper-parameters: number of convolutional and fully connected layers, number of filters per layer, kernel sizes, and feature map normalisation schemes. The CNN architecture providing better performance in our experiments is shown in Figure~\ref{fig:simple_cnn}. Our CNN consists in three convolutional+pooling stages followed by an extra convolution and three fully connected layers. Details about the specific configuration and parameters are given in section~\ref{sec:implementation}.

At testing time, given a query scene text image the trained CNN model is applied to image patches following the same sampling strategy described before. Then, the individual CNN responses for each image patch can be fed into the global classification rule in order to make a single labeling decision for the query image.

\subsection{Training with an Ensemble of Conjoined Networks}

Since the output of the CNN for an individual image patch is a probability distribution over class labels, a simple global decision rule would be just to average the responses of the CNN for all patches in a given query image:

\begin{equation}
\label{eq_simple}
y^{(I)} = \frac{1}{n_I}\sum_{i=1}^{n_I} CNN(x_i)
\end{equation}

where an image $I$ takes the label with more probability in the averaged softmax responses ($y^{(I)}$) of their $n_I$ individual patches $\{x_1, ... , x_{n_I}\}$ outputs on the CNN. \\

The problem with this global classification rule is that the CNN weights have been trained to solve a problem (individual patch classification) that is different from the final goal (i.e. classifying the whole query image). Besides, it is based in a simplistic voting strategy for which all patches are assumed to weight equally, i.e. no patches are more or less discriminative than others. To overcome this we propose the use of an Ensemble of Conjoined Nets in order to train the CNN for a task that resembles more the final classification goal. 

\begin{figure*}[h]
\centering
\begin{tikzpicture}[scale=0.80]
\input{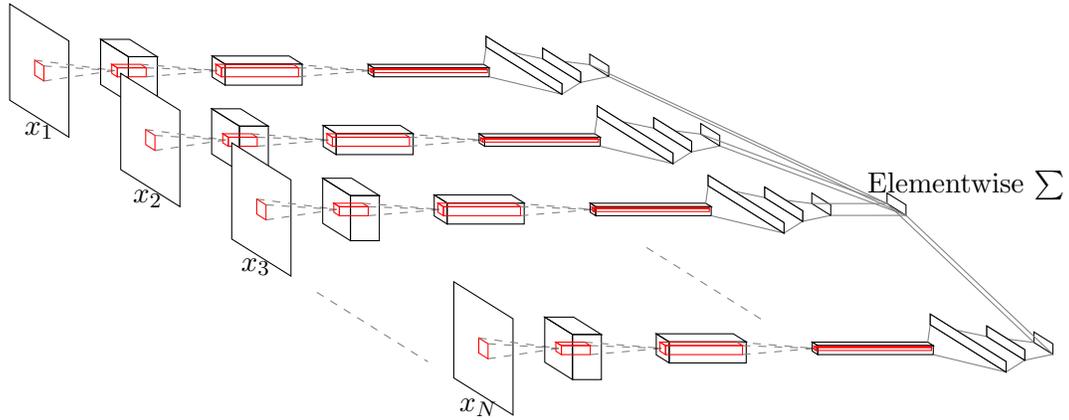}
\end{tikzpicture}
\caption{An Ensemble of Conjoined Nets consist in a set of identical networks that are joined at their outputs in order to provide a unique classification response.}
\label{fig:ecn_cnn}
\end{figure*}

An Ensemble of Conjoined Nets (ECN), depicted in Figure~\ref{fig:ecn_cnn}, consists in a set of identical networks that are joined at their outputs in order to provide a unique classification response. At training time the ECN is presented with a set of $N$ image patches extracted from the same image, thus sharing the same label, and produces a single output for all them. Thus, to train an ECN we must build a new training dataset where each sample consists in a set of $N$ patches with the same label (extracted from the same image).

ECNs take inspiration from Siamese Networks~\cite{Bromley1993} but, instead of trying to learn a metric space with a distance-based loss function, the individual networks in the ECN are joined at their last fully connected layer (fc7 in our case), which has the same number of neurons as the number of classes, with a simple element-wise sum operation and thus we can use the standard cross-entropy classification loss. This way, the cross-entropy classification loss function of the ECN can be written in terms of the $N$ individual patch responses as follows:

\begin{equation}
\label{eq_loss}
\begin{split}
E = \frac{-1}{M} \sum\limits_{m=1}^M \log( \hat{p}_{m,l_m} ),\\
\hat{p}_{m,k} = \exp(\sum\limits_{n=1}^N x_{mnk}) / \left[\sum\limits_{k'=1}^K \exp(\sum\limits_{n=1}^N x_{mnk'})\right]
\end{split}
\end{equation}

where $M$ is the number of input samples in a mini-batch, $\hat{p}_{m}$ is the probability distribution over classes provided by the softmax function, $l_m$ is the label of the $m$'th sample, $N$ is the number of conjoined networks in the ensemble, $K$ is the number of classes, and $x_{mnk} \in [-\infty, +\infty] $ indicates the response (score) of the $k$'th neuron in the $n$'th network for the $m$'th sample.

As can be appreciated in equation~\ref{eq_loss}, in an ECN network a single input patch contributes to the backpropagation error in terms of a global goal function for which it is not the only patch responsible. For example, even when a single patch is correctly scored in the last fully connected layer it may be penalized, and induced to produce a larger activation, if the other patches in its same sample contribute to a wrong classification at the ensemble output.

At test time, the CNN model trained in this way is applied to all image patches in the query image and the global classification rule is defined as:

\begin{equation}
\label{eq_simple_fc7}
y^{(I)} = \sum_{i=1}^{n_I} CNN_{fc7}(x_i)
\end{equation}

where an image $I$ takes the label with the highest score in the sum ($y^{(I)}$) of the fc7 layer responses of the $n_I$ individual patches $\{x_1, ... , x_{n_I}\}$. This is the same as in Equation~\ref{eq_simple} but using the fc7 layer responses instead of the output softmax responses of the CNN.

Notice that still the task for which the ECN network has been trained is not exactly the same defined by this global classification rule, as the number of patches $n_I$ is variable for each image and usually different than the number of conjoined networks $N$. However, it certainly resembles more the true final classification goal. The number of conjoined networks $N$ is an hyper-parameter of the method that is largely dependent on the task to be solved and is discussed in the experimental section.

%TODO add explanation on how the ECN is a kind of data augmentation ... discussionn

%TODO show in an image the random N-combinations of patches in Figure 3

%------------------------------------------------------------------------- 
\section{Experiments}
\label{sec:experiments}
All reported experiments were conducted over three datasets, namely the Video Script Identification Competition (CVSI-2015) dataset\footnote{\url{http://www.ict.griffith.edu.au/cvsi2015/}}, the SIW-13 dataset\footnote{\url{http://mc.eistar.net/~xbai/mspnProjectPage/}}, and the MLe2e dataset\footnote{\url{http://github.com/lluisgomez/script_identification/}}.

The CVSI-2015~\cite{Sharmai2015} dataset comprises pre-segmented words in ten scripts: English, Hindi, Bengali, Oriya, Gujrathi, Punjabi, Kannada, Tamil, Telegu, and Arabic. The dataset contains about 1000 words for each script and is divided into three parts: a training set ($~60\%$ of the total images), a validation set ($10\%$), and a test set ($30\%$). Text is extracted from various video sources (news, sports etc.) and, while it contains a few instances of scene text, it covers mainly overlay video text.

The SIW-13 datset~\cite{Shi2016} comprises $16291$ pre-segmented text lines in thirteen scripts: Arabic, Cambodian, Chinese, English, Greek, Hebrew, Japanese, Kannada, Korean, Mongolian, Russian, Thai, and Tibetan. The test set contains $500$ text lines for each script, $6500$ in total, and all the other images are provided for training. In this case, text was extracted from natural scene images from Google Street View. 

\subsection{The MLe2e dataset}

This paper introduces the first dataset available up to date for the evaluation of multi-lingual scene text end-to-end reading systems and all intermediate stages: text detection, script identification, and text recognition. The Multi-Language end-to-end (MLe2e) dataset has been harvested from various existing scene text datasets for which the images and ground-truth have been revised in order to make them homogeneous. The original images come from the following datasets: Multilanguage(ML)~\cite{Pan2009} and MSRA-TD500~\cite{yao2012detecting} contribute Latin and Chinese text samples, Chars74K~\cite{deCampos09} and MSRRC~\cite{kumar2013multi} contribute Latin and Kannada samples, and KAIST~\cite{lee2010} contributes Latin and Hangul samples.

In order to provide a homogeneous dataset, all images have been resized proportionally to fit in $640\times480$ pixels, which is the default image size of the KAIST dataset. Moreover, the groundtruth has been revised to ensure a common text line annotation level~\cite{karatzas2014line}. %Notice that since in Chars74K, MSRRC and KAIST the original annotations are at the word level, this has required a significant amount of manual work. 
During this process human annotators were asked to review all resized images, adding the script class labels and text transcriptions to the groundtruth, and checking for annotation consistency: discarding images with unknown scripts or where all text is unreadable (this may happen because images were resized); joining individual word annotations into text line level annotations; discarding images where correct text line segmentation is not clear or cannot be established, and images where a bounding box annotation contains more than one script (this happens very rarely e.g. in trademarks or logos) or where more than half of the bounding box is background (this may happen with heavily slanted or curved tex). Arabic numerals ($0,..,9$), widely used in combination with many (if not all) scripts, are labeled as follows. A text line containing text and Arabic numerals is labeled as the script of the text it contains, while a text line containing only Arabic numerals is labeled as Latin. 

The MLe2e dataset contains a total of $711$ scene images covering four different scripts (Latin, Chinese, Kannada, and Hangul) and a large variability of scene text samples. The dataset is split into a train and a test set with $450$ and $261$ images respectively. The split was done randomly, but in a way that the test set contains a balanced number of instances of each class (aprox. $160$ text lines samples of each script), leaving the rest of the images for the train set (which is not balanced by default). The dataset is suitable for evaluating various typical stages of end-to-end pipelines, such as multi-script text detection, joint detection and script identification, end-to-end multi-lingual recognition, and script identification in pre-segmented text lines. For the latter, the dataset also provides the cropped images with the text lines corresponding to each data split: $1178$ and $643$ images in the train and test set respectively.

While being a dataset that has been harvested from a mix of existing datasets it is important to notice that building it has supposed an important annotation effort: since some of the original datasets did not provide text transcriptions, and/or where annotated at different granularity levels. Moreover, despite the fact that the number of languages in the dataset is rather limited (four scripts) it is the first public dataset that covers the evaluation of all stages of multi-lingual end-to-end systems for scene text understanding in natural scenes. We think this is an important contribution of this paper and hope the dataset will be useful to other researchers in the community.

\subsection{Implementation details}
\label{sec:implementation}

In this section we detail the architectures of the network models used in this paper, as well as the different hyper-parameter setups that can be used to reproduce the results provided in following sections.  In all our experiments we have used the open source Caffe~\cite{jia2014caffe} framework for deep learning running on commodity GPUs. Source code and compatible Caffe models are made publicly available\footnote{\url{http://github.com/lluisgomez/script_identification/}}.

We have performed exhaustive experiments by varying many of the proposed method’s parameters, training multiple models, and choosing the one with best cross-validation performance on the SIW-13 training set. The following parameters were tuned in this procedure: the  size and step of the sliding window, the base learning rate, the number of convolutional and fully connected layers, the number of nodes in all layers, the convolutional kernel sizes, and the feature map normalisation schemes

This way, the best basic CNN model found for individual image patch classification is described in section~\ref{sec:simple_cnn} and Figure~\ref{fig:simple_cnn}, and has the following per layer configuration:

\begin{itemize}
\item{Input layer: single channel $32\times32$ image patch.}
\item{conv1 layer: $96$ filters with size $5\times5$. Stride=1, pad=0. Output size: $96\times28\times28$.}
\item{pool1 layer: kernel size=3, stride=2, pad=1. Otput size: $96\times15\times15$.}
\item{conv2 layer: 256 filters with size $3\times3$. Stride=1, pad=0. Output size: $256\times13\times13$.}
\item{pool2 layer: kernel size=3, stride=2, pad=1. Otput size: $256\times7\times7$.}
\item{conv3 layer: 384 filters with size $3\times3$. Stride=1, pad=0. Output size: $384\times5\times5$.}
\item{pool3 layer: kernel size=3, stride=2, pad=1. Otput size: $384\times3\times3$.}
\item{conv4 layer: 512 filters with size $1\times1$. Stride=1, pad=0. Output size: $512\times3\times3$.}
\item{fc5 layer: 4096 neurons.}
\item{fc6 layer: 1024 neurons.}
\item{fc7 layer: $N$ neurons, where $N$ is the number of classes.}
\item{SoftMax layer: Output a probability distribution over the $N$ class labels.}
\end{itemize}

The total number of parameters of the network is $\approx24$M for the $N=13$ case in the SIW-13 dataset. 
All convolution and fully connected layers use Rectified Linear Units (ReLU). In conv1 and conv2 layers we perform normalization over input regions using Local Response Normalization (LRN)~\cite{jarrett2009}. At training time, we use dropout~\cite{srivastava2014dropout} (with a $0.5$ ratio) in fc5 and fc6 layers. 

To train the basic network model we use Stochastic Gradient Descent (SGD) with momentum and $L2$ regularization. We use mini-batches of $64$ images. The base learning rate is set to $0.01$ and is decreased by a factor of $\times10$ every $100$k iterations. The momentum weight parameter is set to $0.9$, and the weight decay regularization parameter to $5\times10^{-4}$. 

When training for individual patch classification, we build a dataset of small patches extracted by dense sampling the original training set images, as explained in section~\ref{sec:simple_cnn}. Notice that this produces a large set of patch samples, e.g. in the SIW-13 dataset the number of training samples is close to half million. With these numbers the network converges after $250$k iterations.

In the case of the Ensemble of Conjoined Networks the basic network detailed above is replicated $N$ times, and all replicas are tied at their fc7 outputs with an element-wise sum layer which is connected to a single output SoftMax layer. All networks in the ECN share the same parameters values.

Training the ECN requires a dataset where each input sample is composed by $N$ image patches. We generate this dataset as follows: given an input image we extract patches the same way as for the simple network, then we generate random $N-$combinations of the image patches, allowing repetitions if the number of patches is $<N$. Notice that this way the number of samples can be increased up to very large-scale numbers because the number of possible different $N-$combinations is $\tbinom MN$ when the number of patches in a given image $M$ is larger that the number of conjoined nets $N$, which is the usual case. This is an important aspect of ECNs, as the training dataset generation process becomes a data augmentation technique in itself. We can see this data augmentation process as generating new small text instances that are composed from randomly chosen parts of their original generators. 
% For datasets without a validation set we can use also this generation technique to provide one, allowing us to use early-stopping ... 

However, it is obviously non-practical to use all possible combinations for training; thus, in order to get a manageable number of samples, we have used the simple rule of generating $2 \times M$ samples per input, which for example in the SIW-13 dataset would produce around one million samples.

In terms of computational training complexity, the ECN has an important drawback compared to the simple network model: the number of computations is multiplied by $N$ in each forward pass, similarly the amount of memory needed is linearly increased by $N$. To overcome this limitation, we use a fine-tuning approach to train ECNs. First, we train the simple network model, and then we do fine-tuning on the ECN parameters starting from the values learned using the simple net. When fine-tuning, we have found that starting from a fully converged network in the single-patch classification task we reach a local minimum of the global task, thus providing zero loss in most (if not all) the iterations and not allowing the network to learn anything new. %What worked the best for us 
In order to avoid this local minima situation we start the fine-tuning from a non-converged network (more or less at about 90/95\% of the attainable individual patch classification accuracy).

Using fine-tuning with a base learning rate of $0.001$ (decreasing $\times10$ every $10$k iterations) the ECN converges much faster, in the order of $35$k iterations. All other learning parameters are set the same as in the simple network training setup.

\begin{figure}[h]
\centering
\includegraphics[width=0.5\linewidth]{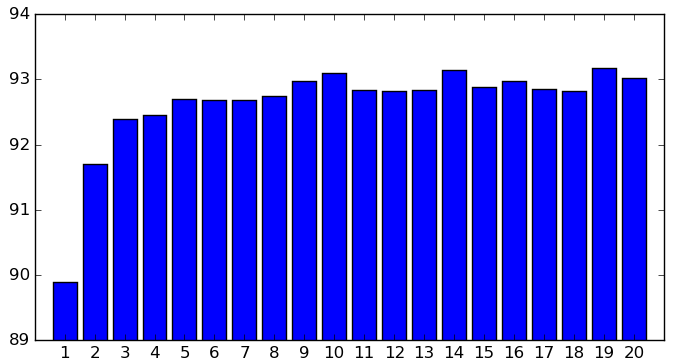}
\caption{Validation accuracy for various number of networks $N$ in the ensemble of conjoined networks model.}
\label{fig:ECN_optimal_N}
\end{figure}

The number of nets $N$ in the ensemble can be seen as an extra hyper-parameter in the ECN learning algorithm. Intuitively a dataset with larger text sequences would benefit from larger $N$ values, while on the contrary in the extreme case of classifying small squared images (i.e. each image is represented by a single patch) any value of $N>1$ does not make sense. Since our datasets contain text instances with variable length a possible procedure to select the optimal value of $N$ is by using a validation set. We have done experiments in the SIW-13 dataset by dividing the provided train set and keeping $10\%$ for validation. Classification accuracy on the validation set for various $N$ values are shown in Figure~\ref{fig:ECN_optimal_N}. As can be appreciated the positive impact of training with an ensemble of networks is evident for small values of $N$, and mostly saturated for values $N>9$. In the following we use a value of $N=10$ for all the remaining experiments.

\subsection{Script identification in pre-segmented text lines}
\label{sec:exp_ident}

\begin{figure}[b]
\includegraphics[width=0.32\linewidth]{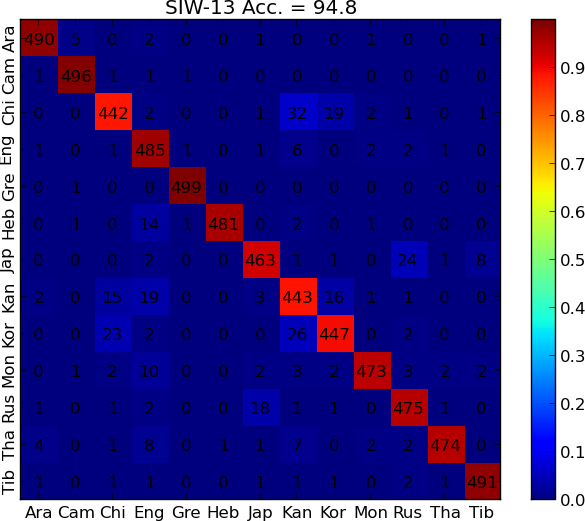}
\includegraphics[width=0.32\linewidth]{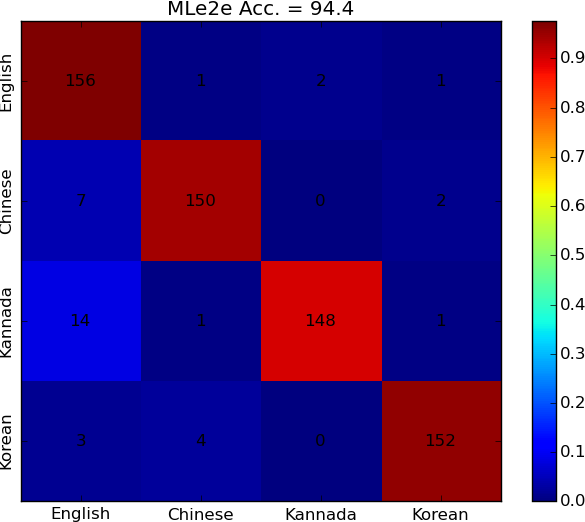}
\includegraphics[width=0.32\linewidth]{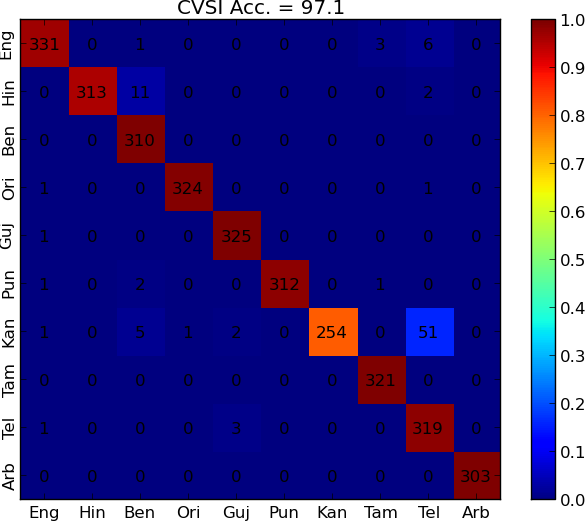}
\caption{Confusion matrices with per class classification accuracy of our method in SIW-13, MLe2e, and CVSI datasets.}
\label{fig:confusion}
\end{figure}

In this section we study the performance of the proposed method for script identification in pre-segmented text lines. Table~\ref{tab:cvsi} shows the overall performance comparison of our method with the state-of-the-art in CVSI-2015, SIW-13, and MLe2e datasets. Figure~\ref{fig:confusion} shows the confusion matrices for our method in all three datasets with detailed per class classification results. 

In Table~\ref{tab:cvsi} we also provide comparison with three well known image recognition pipelines using Scale Invariant Features~\cite{lowe1999object} (SIFT) in three different encodings: Fisher Vectors, Vector of Locally Aggregated Descriptors (VLAD), and Bag of Words (BoW); and a linear SVM classifier. In all baselines we extract SIFT features at four different scales in sliding window with a step of 8 pixels. For the Fisher vectors we use a 256 visual words GMM, for VLAD a 256 vector quantized visual words, and for BoW 2,048 vector quantized visual words histograms. The step size and number of visual words were set to similar values to our method when possible in order to offer a fair evaluation. These three pipelines have been implemented with the VLFeat~\cite{vedaldi08vlfeat} and liblinear~\cite{liblinear} open source libraries. The entry ``Sequence-based CNN'' in Table~\ref{tab:cvsi} corresponds to the results obtained with the early fusion design proposed in~\cite{karpathy2014large} with a stack of 5 consecutive patches.

%%TODO add baseline methods from Shi et al 2015/16?
\begin{table}
\begin{center}
\small
\begin{tabularx}{\textwidth}{ X c c c c c }
\hline
\toprule
Method & SIW-13 & MLe2e & CVSI\\
\midrule
\textbf{This work - Ensemble of Conjoined Nets } & \textbf{94.8} & \textbf{94.4} & 97.2\\
\textbf{This work - Simple CNN (Avg.)} & 92.8 & 93.1 & 96.7\\
\textbf{This work - Simple CNN (fc5+SVM)} & 93.4 & 93.6 & 96.9\\
\midrule
Shi~\etal~\cite{Shi2016} & 89.4 & - & 94.3\\
HUST~\cite{Shi2015,Sharmai2015} & 88.0 & - & 96.69\\
Google~\cite{Sharmai2015} & - & - & \textbf{98.91} \\
Nicolaou~\etal~\cite{Nicolaou2016} & 83.7 & - & 98.18\\
Gomez~\etal~\cite{Gomez2016} & 76.9 & 91.12 & 97.91\\
CVC-2~\cite{Gomez2016,Sharmai2015} & - & 88.16 & 96.0\\
SRS-LBP + KNN~\cite{nicolaou2015sparse} & - & 82.71 & 94.20\\
%CVC-1 & 95.88 & - & - & -\\
C-DAC~\cite{Sharmai2015} & - & - & 84.66\\
CUK~\cite{Sharmai2015} & - & - & 74.06\\
\midrule
Baseline SIFT + Fisher Vectors + SVM & 90.7 & 88.63 & 94.11 \\
Baseline SIFT + VLAD + SVM & 89.2 & 90.19 & 93.92 \\
Baseline SIFT + Bag of Words + SVM & 83.4 & 86.45 & 84.38 \\
Baseline Sequence-based CNN~\cite{karpathy2014large} (Early fusion) & 88.9 & 89.80 & 93.62 \\
\bottomrule
\end{tabularx}
\end{center}
\caption{Overall classification performance comparison with state-of-the-art in three different datasets: SIW-13~\cite{Shi2016}, MLe2e, and CVSI~\cite{Sharmai2015}.}
\label{tab:cvsi}
\end{table}

As shown in Table~\ref{tab:cvsi} the proposed method outperforms state of the art and all baseline methods in the SIW-13 and MLe2e scene text datasets, while performing competitively in the case of CVSI video overlay text dataset. In the SIW-13 dataset the proposed method significantly outperforms the best performing method known up to date by more than $4$ percentual points.

The entry ``Simple CNN (fc5+SVM)'' (third row) in Table~\ref{tab:cvsi} corresponds to the results obtained with a linear SVM classifier by using features extracted from the ``Simple CNN'' network. For this experiment we represent each image in the dataset with a fixed length vector with the averaged outputs of all its patches in the fc5 layer of the network. Then we train a linear SVM classifier using cross-validation on the training set, and show the classification performance on the test set. Similar results (or slightly worse) have been found for features extracted from other layers (fc7, fc6, conv4) and using other linear classifiers (e.g. logistic regression). When compared with the ``Simple CNN'' approach we appreciate that classification performance is better for this combination (fc5+SVM). This confirms the intuition that classification performance can be improved by optimizing the combination of the results for the individual patches. However, the performance of the CNN trained with the ensemble of conjoined networks is still better. As mentioned earlier, the additional benefit of our approach here is in the end-to-end learning of both the visual features and the optimal combination scheme for classification.

The contribution of training with ensembles of conjoined nets is consistent in all three evaluated datasets but more notable on SIW-13, as appreciated by comparing the first two rows of Table~\ref{tab:cvsi} which correspond to the nets trained with the ensemble (first row) and the simple model (second row). This comparison can be further strengthened by testing if the provided improvement is statistically significant. For this we use the within-subjects chi-squared test (McNemar's test)~\cite{mcnemar1947note} to compare the predictive accuracy of the two models. The obtained p-values on the SIW-13, MLe2e, and CVSI datasets are respectively $1.4\times10^{-16}$, $0.057$, and $0.0026$. In the case of the SIW-13 dataset the p-value is way smaller than the assumed significance threshold ($\alpha=0.05$), thus we can reject the null-hypothesis that both models perform equally well on this dataset and certify a statistically significant improvement. On the other hand we appreciate a marginal improvement on the other two datasets.

%TODO would be interesting to discuss why we are better than Yao holistic net (this is about Joost' scepticism) ... and why it would be ineteresting to have a public implementation of yao to make a fair comparison

\begin{figure}[b]
\includegraphics[width=\linewidth]{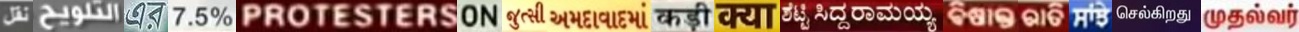}\\
\includegraphics[width=\linewidth]{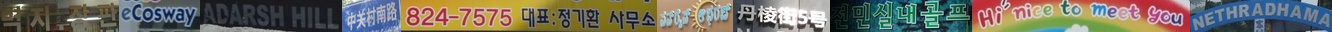}
\caption{Overlaid-text samples (top row) variability and clutter is rather limited compared with that found in the scene text images (bottom row).}
\label{fig:dataset_diff}
\end{figure}

\begin{figure}[t]
\includegraphics[width=\linewidth]{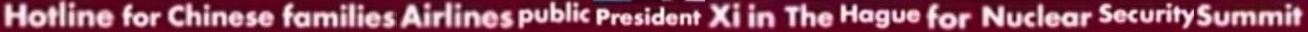}\\
\includegraphics[width=\linewidth]{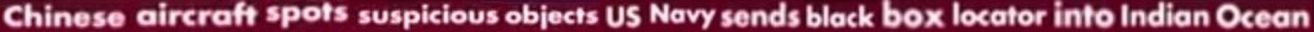}
\caption{Cropped words in the CVSI dataset belong to very long sentences of overlay text in videos. It is common to find several samples sharing exactly the same font and background both in the train (top row) and test (bottom row) sets.}
\label{fig:cvsi_similar}
\end{figure}

Our interpretation of the results on CVSI and MLe2e datasets in comparison with the ones obtained on SIW-13 relates to its distinct nature. On one hand the MLe2e dataset covers only four different scripts (Latin, Chinese, Kannada, and Hangul) for which the inter-class similarity does not represent a real problem. On the other hand, the CVSI overlaid-text variability and clutter is rather limited compared with that found in the scene text of MLe2e and SIW-13. As can be appreciated in Figure~\ref{fig:dataset_diff} overlaid-text is usually bi-level without much clutter. Figure~\ref{fig:cvsi_similar} shows another important characteristic of CVSI dataset: since cropped words in the dataset belong to very long sentences of overlay text in videos, e.g. from rotating headlines, it is common to find a few dozens of samples sharing exactly the same font and background both in the train and test sets. This particularity makes the ECN network not really helpful in the case of CVSI, as the data augmentation by image patches recombinations is somehow already implicit on the dataset.

Furthermore, the CVSI-2015 competition winner (Google) makes use of a deep convolutional network but applies a binarization pre-processing to the input images. In our opinion this binarization may not be a realistic preprocessing in general for scene text images. As an example of this argument one can easily see in Figure~\ref{fig:dataset_diff} that binarization of scene text instances is not trivial as in overlay text. Similar justification applies to other methods performing better than ours in CVSI. In particular the LBP features used in~\cite{Nicolaou2016}, as well as the patch-level whitening used in our previous work~\cite{Gomez2016}, may potentially take advantage of the simpler, bi-level, nature of text instances in CVSI dataset. It is important to notice here that these two algorithms, corresponding to our previous works in script identification, have close numbers to the Google ones in CVSI-2015 (see Table~\ref{tab:cvsi}) but perform quite bad in SIW-13.

\begin{figure}[t]
\includegraphics[height=3.5em]{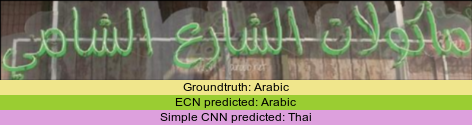} \includegraphics[height=3.5em]{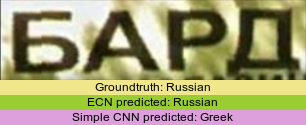} \includegraphics[height=3.5em]{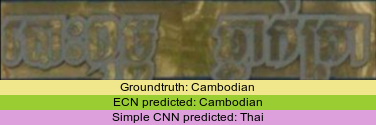}\\
\includegraphics[height=3.5em]{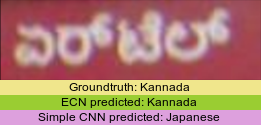} \includegraphics[height=3.5em]{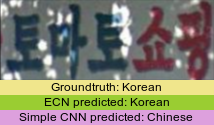} \includegraphics[height=3.5em]{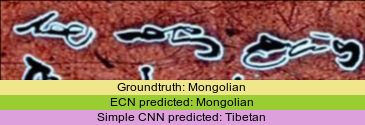} \includegraphics[height=3.5em]{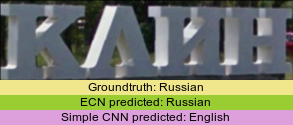}
\caption{Examples of challenging text images that are correctly classified by our ECN method but not with the Simple CNN approach.}
\label{fig:img_improvements}
\end{figure}

\begin{figure}[t]
\includegraphics[width=\linewidth]{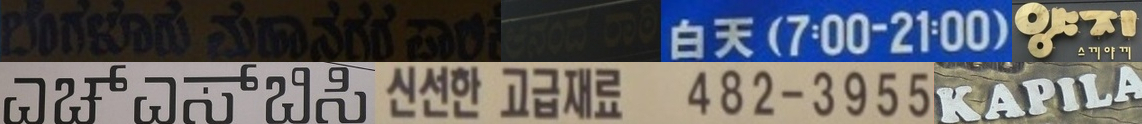}
\caption{A selection of %16 of the 57 
misclassified samples by our method: low contrast images, rare font types, degraded text, letters mixed with numerals, etc.}
\label{fig:babel_errors}
\end{figure}

As a conclusion of the experiments performed in this section we can say that the improvement of training a patch-based CNN classifier with an ensemble of conjoined nets is especially appreciable in cases where we have a large number of classes, with large inter-class similarity, and cluttered scene images, as is the case of the challenging SIW-13 dataset. This demonstrates our initial claim that a powerful script identification method for scene text images must be based in learning good local patch representations, and also their relative importance in the global image classification task. Figure~\ref{fig:img_improvements} shows some examples of challenging text images that are correctly classified by our method but not with the Simple CNN approach. Figure~\ref{fig:babel_errors} shows a set of missclassified images.

\begin{figure}[t]
\includegraphics[width=\linewidth]{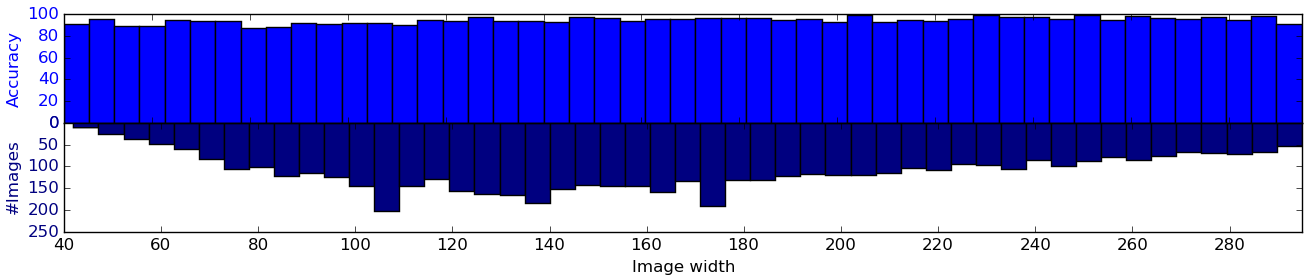}
\caption{Classification accuracy of the CNN trained with ensembles of conjoined nets (top) and number of images (bottom) as a function of the image width on SIW-13 test images.}
\label{fig:acc_vs_sidth}
\end{figure}

Finally, in Figure~\ref{fig:acc_vs_sidth} we show the classification accuracy of the CNN trained with ensembles of conjoined nets as a function of the image width on SIW-13 test images. We appreciate that the method is robust even for small text images which contain a limited number of unique patches. Computation time for our method is also dependent on the input image length and ranges from $4$ms. in for the smaller images up to $23$ms. for the larger ones. The average computation time on the SIW-13 test set is of $13$ms using a commodity GPU. At test time computation is made efficient by stacking all patches of the input image in a single mini-batch.

\subsection{Joint text detection and script identification in scene images}

In this experiment we evaluate the performance of a complete pipeline for detection and script identification in its joint ability to detect text lines in natural scene images and properly recognizing their scripts. The key interest of this experiment is to study the performance of the proposed script identification algorithm when realistic, non-perfect, text localisation is available.

Most text detection pipelines are trained explicitly for a specific script (typically English) and generalise pretty badly to the multi-script scenario. We have chosen to use here our previously published script-agnostic method~\cite{Gomez2014}, which is designed for multi-script text detection and generalises well to any script.
The method detects character candidates using the Maximally Stable Extremal Regions (MSER)~\cite{Matas2004} algorithm, and builds different hierarchies where the initial regions are grouped by agglomerative clustering, using complementary similarity measures. In such hierarchies each node defines a possible text hypothesis. Then, an efficient classifier, using incrementally computable descriptors, is used to walk each hierarchy and select the nodes with larger text-likelihood.

In this paper script identification is performed at the text line level, because segmentation into words is largely script-dependent, and not meaningful in Chinese/Korean scripts. Notice however that in some cases, by the intrinsic nature of scene text, a text line provided by the text detection module may correspond to a single word, so we must deal with a large variability in the length of provided text lines. %paragraph level is not an option since in scene images test is many times appears isolated words, or text lines.
The experiments are performed over the new MLe2e dataset.

%First, we evaluate the text detection performance of \cite{Gomez2014} in the new dataset. Notice that since in our method the detection and recognition modules are independent, i.e. there is no way from the script identification module to recover from errors of the text detection module, the detection performance is an upper bound of the performance in the joint task. The detection module achieved an f-score of $0.53$ using the Deteval~\cite{wolf2006} evaluation framework, and of $0.57$ using the ICDAR2003~\cite{Lucas2003} framework, which is similar to the performance reported in~\cite{Gomez2014} for the MSRRC dataset.

For evaluation of the joint text detection and script identification task in the MLe2e dataset we propose the use of a simple two-stage evaluation framework. First, localisation is assessed based on the Intersection-over-Union (IoU) metric between detected and ground-truth regions, as commonly used in object detection tasks~\cite{everingham2014pascal} and the recent ICDAR 2015 Robust Reading Competition\footnote{\url{http://rrc.cvc.uab.es}}~\cite{karatzas2015}. Second, the predicted script is verified against the ground-truth. A detected bounding box is thus considered correct if it has a IoU $>0.5$ with a bounding box in the ground-truth and the predicted script is correct.

The localisation-only performance, corresponding to the first stage of the evaluation, yields an F-score of $0.63$ (Precision of $0.57$ and Recall of $0.69$). This defines the upper-bound for the joint task. The two stage evaluation, including script identification, of the proposed method compared with our previous work is shown in Table~\ref{tab:joint_detident}.

\begin{table}
\begin{center}
\small
\begin{tabularx}{\textwidth}{ X c c c c c c}
\toprule
Method & Correct & Wrong & Missing & Precision & Recall & F-score\\
\midrule
\textbf{This work - ECN} & \textbf{395} & \textbf{376} & \textbf{245} & \textbf{0.51} & \textbf{0.62} & \textbf{0.56}\\
%\textbf{This work - CNN (Avg.)} & - & - & - & - & - & -\\
Gomez~\etal~\cite{Gomez2016} & 364 &407 & 278 & 0.47 &0.57 & 0.52\\
\bottomrule
\end{tabularx}
\end{center}
\caption{Text detection and script identification performance in the MLe2e dataset.}
\label{tab:joint_detident}
\end{table}

\begin{figure}[b]
\includegraphics[width=0.9\linewidth]{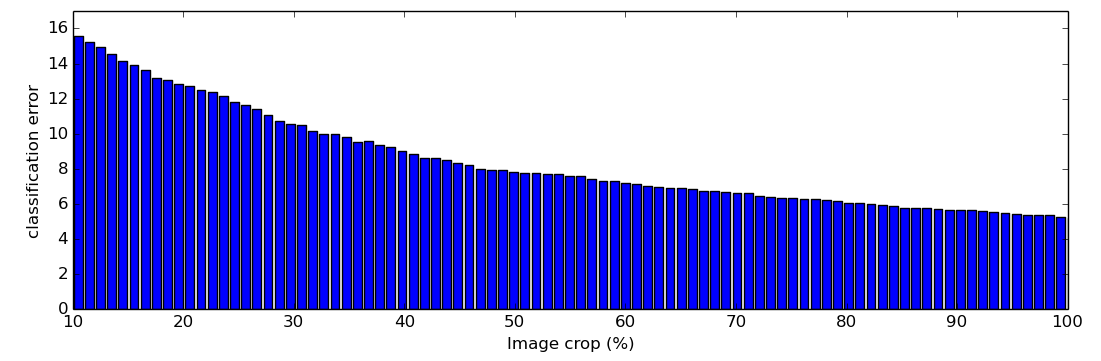}
\caption{Classification error of our method when applied to variable length cropped regions of SIW-13 images, up to the minimum size possible ($40\times40$ pixels).}
\label{fig:acc_crop}
\end{figure}

Intuitively the proposed method for script identification is effective even when the text region is badly localised, as long as part of the text area is within the localised region. To support this argument we have performed an additional experiment where our algorithm is applied to cropped regions from pre-segmented text images. For this, we take the SIW-13 original images and calculate the performance of our method when applied to  cropped regions of variable length, up to the minimum size possible ($40\times40$ pixels). As can be appreciated in Figure~\ref{fig:acc_crop} the experiment demonstrates that the proposed method is effective even when small parts of the text lines are provided. Such a behaviour is to be expected, due to the way our method treats local information to decide on a script class. In the case of the pipeline for joint detection and script identification, this extends to regions that did not pass the $0.5$ IoU threshold, but had their script correctly identified. This opens the possibility to make use of script identification to inform and / or improve the text localisation process. The information of the identified script can be used to refine the detections.

\subsection{End-to-end multi-lingual recognition in scene images}

In this section we evaluate the performance of a complete pipeline for end-to-end multi-lingual recognition in scene images. For this, we combine the pipeline used in the previous section with a well known off-the-shelf OCR engine: the open source project Tesseract\footnote{\url{http://code.google.com/p/tesseract-ocr/}}~\cite{ray2007}. Similar pipelines~\cite{milyaev2013,gomez2014scene,milyaev2015} using off-the-shelf OCR engines have demonstrated state-of-the-art end-to-end performance in English-only datasets up to very recently, provided that the text detection module is able to produce good pixel-level segmentation of text.

The setup of the OCR engine in our pipeline is minimal: given a text detection hypothesis from the detection module we set the recognition language to the one provided by the script identification module, and we set the OCR to interpret the input as a single text line. Apart from that we use the default Tesseract parameters. 

The recognition output is filtered with a simple post-processing junk filter in order to eliminate garbage recognitions, i.e. sequences of identical characters like "IIii" that may appear as the result of trying to recognize repetitive patterns in the scene. Concretely, we discard the words in which more than half of their characters are recognized as one of "i", "l", "I", or other special characters like: punctuation marks, quotes, exclamation, etc. We also reject those detections for which the recognition confidence provided by the OCR engine is under a certain threshold.

The evaluation protocol is similar to the one used in other end-to-end scene text recognition datasets~\cite{Wang2011,karatzas2015}. Ideally, in end-to-end word recognition, a given output word is considered correct if it overlaps more than $0.5$ with a ground-truth word and all its characters are recognized correctly (case sensitive). However, since in the case of the MLe2e dataset we are evaluating text lines instead of single words, we relax a bit this correctness criteria by allowing the OCR output to to make $\frac{1}{8}$ character level errors. This relaxation is motivated by the fact that for a given test sentence with more than 8 characters (e.g. with two words) having only one character mistake may still produce a partial understanding of the text (e.g. one of the words is correct), and thus must not be penalized the same way as if all characters are wrongly recognized. This way, a given output text line is considered correct if overlaps more than $0.5$ with a ground-truth text line and their edit distance divided by the number of characters of the largest is smaller than $\frac{1}{8}$.

\begin{table}
\begin{center}
\small
\begin{tabularx}{\textwidth}{ X c c c c c c }
\toprule
Script identification & Correct & Wrong & Missing & Precision & Recall & F-score\\
\midrule
\textbf{This work - ECN} & \textbf{96} & 212 & \textbf{503} & 0.31 & \textbf{0.16} & \textbf{0.21}\\
%\textbf{This work - CNN (Avg.)} & - & - & - & - & - & -\\
Gomez~\etal~\cite{Gomez2016} & 82 & 211 & 517 & 0.28 &0.14 & 0.18\\
Tesseract & 50 & \textbf{93} & 549 & \textbf{0.35} & 0.08 & 0.13\\
%Tesseract-2 & 51 & \textbf{93} & 548 & \textbf{0.35} & 0.08 & 0.13\\
\bottomrule
\end{tabularx}
\end{center}
\caption{End-to-end multi-lingual recognition performance in the MLe2e dataset.}
\label{tab:end-to-end}
\end{table}

\begin{figure}
\includegraphics[width=\linewidth]{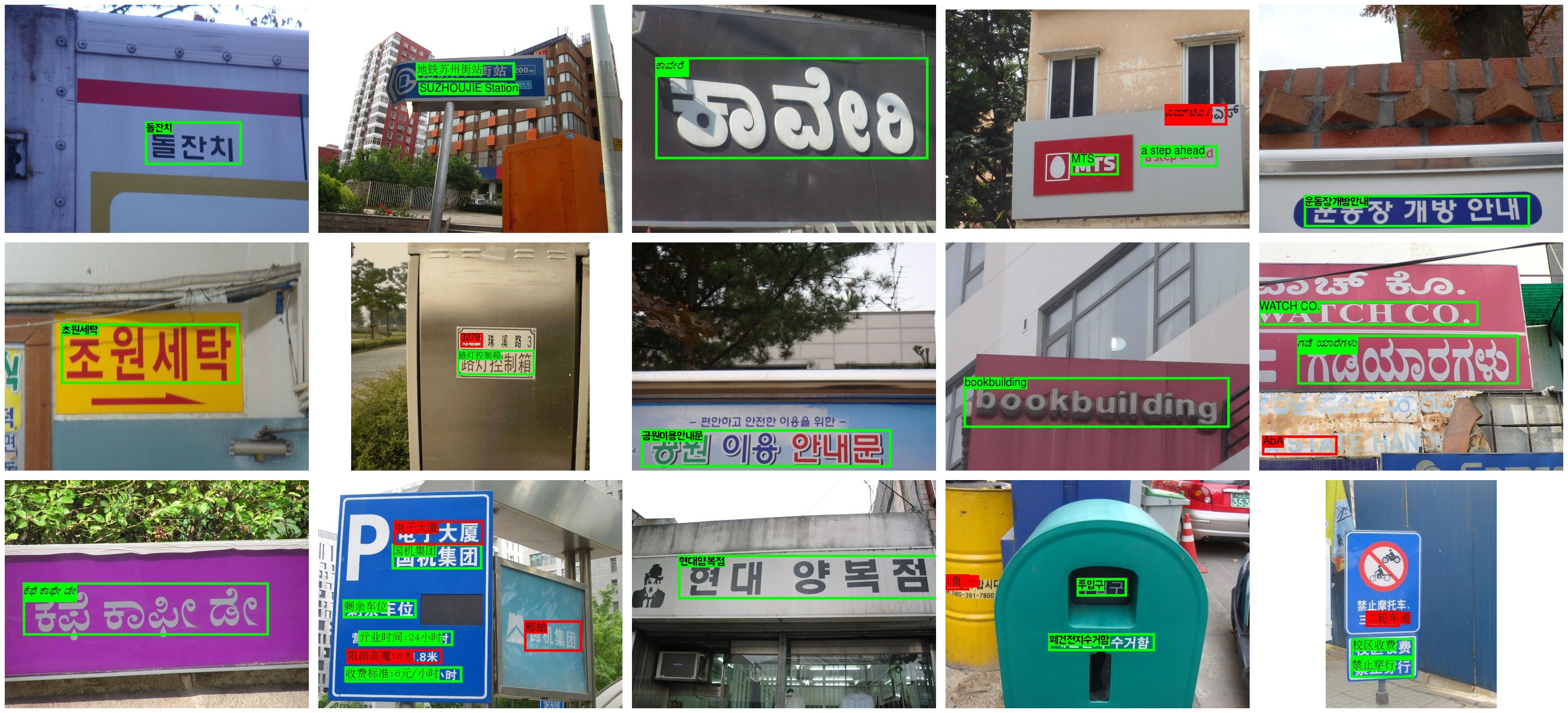}
\caption{End-to-end recognition of text from images containing textual information in different scripts/languages.}
\label{fig:end_results}
\end{figure}

Table~\ref{tab:end-to-end} shows a comparison of the proposed end-to-end pipeline by using different script identification modules: the method presented in this paper, our previously published work, and Tesseract's built-in alternative.  Figure~\ref{fig:end_results} shows the output of our full end-to-end pipeline for some images in the MLe2e test set.

Tesseract method in Table~\ref{tab:end-to-end} refers to the use of Tesseract's own script estimation algorithm~\cite{unnikrishnan2009}. We have found that Tesseract's algorithm is designed to work with large corpses of text (e.g. full page documents) and does not work well for the case of single text lines. 
% and Tesseract-2 refers to the simple procedure of trying to recognize the text in all four possible languages and taking the output with higher OCR confidence. In the first case we have found that Tesseract's algorithm is designed to work with large corpuses of text (e.g. full page documents) and does not work well for the case of single text lines. On the other hand, the Tesseract-2 alternative is provided only as a reference since it obviously deteriorates the recognition time required and would be not a solution in an unrestricted scenario with a large number of possible scripts/languages.

Results in Table~\ref{tab:end-to-end} demonstrate the direct correlation between having better script identification rates and better end-to-end recognition results.

The final multi-lingual recognition f-score obtained ($0.21$) is far from the state-of-the art in end-to-end recognition systems designed for English-only environments~\cite{bissacco2013,Jaderberg2014,Neumann2015}.  As a fair comparison, a very similar pipeline using the Tesseract OCR engine~\cite{gomez2014scene} achieves an f-score of $0.40$ in the ICDAR English-only dataset. The lower performance obtained in MLe2e dataset stems from a number of challenges that are specific to its multi-lingual nature. For example, in some scripts (e.g. Chinese and Kannada) glyphs are many times non single-body regions, composed by (or complemented with) small strokes that in many cases are lost in the text segmentation stage. In such cases having a bad pixel-level segmentation of text would make it practically impossible for the OCR engine to produce a correct recognition.

%TODO compare with recent CNN method for recognition?

Our pipeline results represent the first reference result for multi-lingual scene text recognition and a first benchmark from which better systems can be built, e.g. replacing the off-the-shelf OCR engine by other recognition modules better suited for scene text imagery.

\subsection{Cross-domain performance and confusion in single-language datasets}

In this experiment we evaluate the cross-domain performance of learned CNN weights from one dataset to the other. For example, we evaluate on the MLe2e and CVSI test sets using the network trained with the SIw-13 train set, by measuring classification accuracy only for their common script classes: Arabic, English, and Kannada in CVSI; Chinese, English, Kannada, and Korean in MLe2e. Finally, we evaluate the misclassification error of our method (trained in different datasets) over two single-script datasets. For this experiment we use the ICDAR2013~\cite{karatzas2013icdar} and ALIF~\cite{yousfi2015alif} datasets, which provide cropped word images of English scene text and Arabic video overlaid text respectively. Table~\ref{tab:cross-domain} shows the results of these experiments.

\begin{table}[h]
\begin{center}
\small
\begin{tabularx}{\textwidth}{ X c c c c c }
\toprule
Method 			& SIW-13 & MLe2e & CVSI & ICDAR & ALIF\\
\midrule
ECN CNN (SIW-13) 	& 94.8   & 86.8  & 90.6 & 74.7  & 100 \\
\midrule
ECN CNN (MLe2e) 	& 90.8   & 94.4  & 98.3 & 95.3  & -\\
\midrule
ECN CNN (CVSI) 		& 42.3   & 43.5  & 97.2 & 65.2  & 91.8\\
\bottomrule
\end{tabularx}
\end{center}
\caption{Cross-domain performance of our method measured by training/testing in different datasets.}
\label{tab:cross-domain}
\end{table}

%TODO siw-13 english in icdar is very bad! compare with english class in confusion matrix of siw13

Notice that results in Table~\ref{tab:cross-domain} are not directly comparable among rows because each classifier has been trained with a different number of classes, thus having different rates for a random choice classification. However, the experiment serves as a validation of how good a given classifier is in performing with data that is distinct in nature to the one used for training. In this sense, the obtained results show a clear weakness when the model is trained on the video overlaid text of CVSI and subsequently applied to scene text images (SIW-13, MLe2e, and ICDAR). On the contrary, models trained on scene text datasets are quite stable in other scene text data, as well as in video overlaid text (CVSI and ALIF). 

 In fact, this is an expected result, because the domain of video overlay text can be seen as a sub-domain of the scene text domain. Since the scene text datasets are richer in text variability, e.g. in terms of perspective distortion, physical appearance, variable illumination, and typeface designs, script identification on these datasets is a more difficult problem, and their data is more indicated if one wants to learn effective cross-domain models. This demonstrates that our method is able to learn discriminative stroke-part representations that are not dataset-specific, and provides evidence to the claims made in section~\ref{sec:exp_ident} when interpreting the obtained results in CVSI dataset comparing with other methods that may be more engineered to the specific CVSI data but not generalizing well in scene text datasets.

%\begin{figure}[t]
%\includegraphics[width=\linewidth]{babel_ok.jpg}
%\caption{Examples of correctly classified instances}
%\label{fig:babel_ok}
%\end{figure}

%\begin{figure*}[t]
%\includegraphics[width=\linewidth]{joint_task.jpg}
%\caption{Sample results of our method for the joint task of text detection and script recognition in natural scenes.}
%\label{fig:joint_results}
%\end{figure*}

%------------------------------------------------------------------------- 
\section{Conclusion}
\label{sec:conclusion}
A novel method for script identification in natural scene images was presented. The method is based on the use of ensembles of conjoined convolutional networks to jointly learn discriminative stroke-part representations and their relative importance in a patch-based classification scheme. Experiments performed in three different datasets exhibit state of the art accuracy rates in comparison to a number of state-of-the-art methods, including the participants in the CVSI-2015 competition and three standard image classification pipelines. 

In addition, a new public benchmark dataset for the evaluation of all stages of multi-lingual end-to-end scene text reading systems was introduced.

Our work demonstrates the viability of script identification in natural scene images, paving the road towards true multi-lingual end-to-end scene text understanding.

\section*{Acknowledgment}
\label{sec:acknowledgement}
This project was supported by the Spanish project TIN2014-52072-P, the fellowship RYC-2009-05031, and the Catalan government scholarship 2014FI\_B1-0017.



\section*{References}
  \bibliographystyle{elsarticle-num} 
  \bibliography{GomezKaratzas_pr_2016a}


%\begin{thebibliography}{00}


%\bibitem{}

%\end{thebibliography}
\end{document}